\documentclass{article}
\usepackage{spconf,amsmath,amssymb,graphicx,hyperref}

\title{LiDAR-based Human Activity Recognition \\ through Laplacian Spectral Analysis}
\name{\begin{tabular}{c}Sasan Sharifipour$^{\dagger}$$^{\ast}$, Constantino Álvarez Casado$^{\dagger}$$^{\ddagger}$$^{\ast}$, Le Nguyen$^{\dagger}$$^{\ast}$, \\ Tharindu Ekanayake$^{\dagger}$, Manuel Lage Cañellas$^{\dagger}$, Nhi Nguyen$^{\dagger}$, Miguel Bordallo López$^{\dagger}$\end{tabular}}

\address{
$^{\dagger}$Center for Machine Vision and Signal Analysis (CMVS), University of Oulu, Finland \\
$^{\ddagger}$Candour Ltd, Oulu, Finland \\
\thanks{$^{\ast}$These authors contributed equally to this work.}
}

\begin{document}
% \ninept
%
\maketitle
\begin{abstract}
Human Activity Recognition supports applications in healthcare, manufacturing, and human–machine interaction. LiDAR point clouds offer a privacy-preserving alternative to cameras and are robust to illumination. We propose a HAR method based on graph spectral analysis. Each LiDAR frame is mapped to a proximity graph ($\varepsilon$-graph) and the Laplacian spectrum is computed. Eigenvalues and statistics of eigenvectors form pose descriptors, and temporal statistics over sliding windows yield fixed vectors for classification with Support Vector Machines and Random Forests. In the MM-Fi dataset with 40 subjects and 27 activities, under a strict subject-independent protocol, the method reaches 94.4\% accuracy on a 13-class rehabilitation set and  90.3\% on all 27 activities. It also surpasses the skeleton-based baselines reported for MM-Fi. The contribution is a compact and interpretable feature set derived directly from point-cloud geometry that provides an accurate and efficient alternative to end-to-end deep learning and is applicable to any point cloud source. The code is available at: \url{https://github.com/Arritmic/oulu-pointcloud-har}
\end{abstract}
%
%%%%%%%%%%%%%%%%%%%%%%%%%%%%%%%%%%%%%%%%%%%%%%%%%%%%%%%%%%%%%
%
%
%       INTRODUCTION
%
%
%%%%%%%%%%%%%%%%%%%%%%%%%%%%%%%%%%%%%%%%%%%%%%%%%%%%%%%%%%%%%

\section{Introduction}

% Human Activity Recognition (HAR) supports applications in healthcare, manufacturing, smart environments, and human-machine interaction by allowing systems to detect and classify actions in real time \cite{Rinchi2023IoTMag}. Examples include monitoring rehabilitation exercises, ensuring safety on factory floors, context-aware smart homes, improved sports analytics, or enhancing context-awareness in communication technologies such as 5G or 6G, integrating sensing capabilities into communication networks. Many deployments require methods that respect privacy, operate under changing illumination, and run with limited computational resources, as shown in Figure \ref{fig:pc_person_detection}.

\begin{figure}[ht!]
% \vspace{-3mm}
 \begin{center}
   \includegraphics*[width=0.49\textwidth]{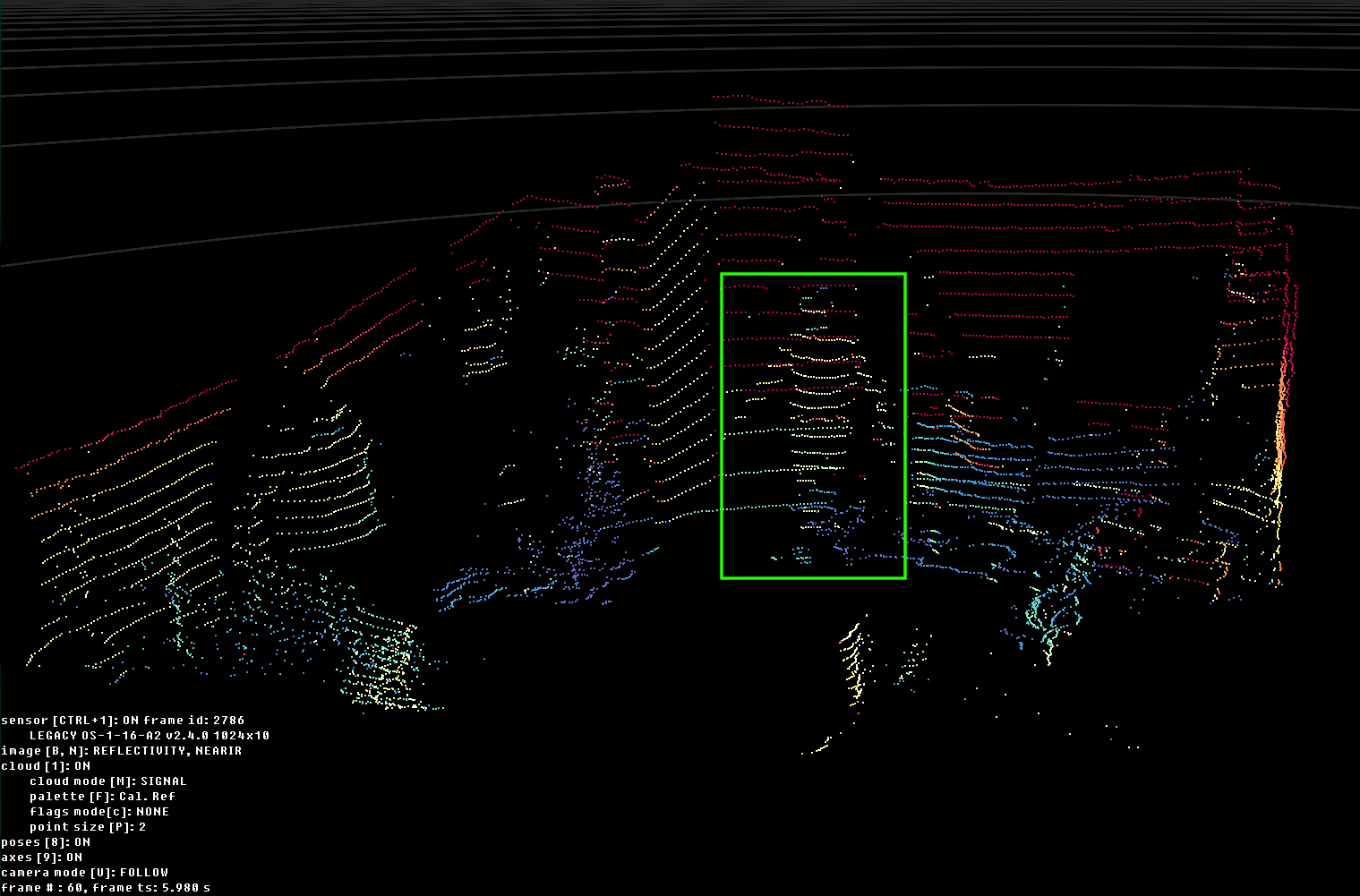}
 \end{center}
 \vspace{-6mm}
 \caption{Indoor 3D point cloud captured with an Ouster OS1 LiDAR at the CMVS Laboratory (University of Oulu). The scene shows a person moving in a room. A bounding box produced by a person detector is overlaid to indicate the detected subject. The frame illustrates the sparse and irregular structure of raw LiDAR data and motivates graph based processing for downstream recognition.}
 \label{fig:pc_person_detection}
 \vspace{-3mm}
\end{figure}

Human Activity Recognition (HAR) is critical for applications in healthcare, manufacturing, and smart environments, many of which require methods that respect privacy, operate under changing illumination, and run on resource-constrained hardware \cite{Rinchi2023IoTMag}. Examples include monitoring rehabilitation exercises, ensuring safety on factory floors, improving sports analytics, or enhancing context-awareness in communication technologies such as 5G or 6G. Common sensing modalities present challenges: cameras raise privacy issues, wearables require user compliance, and radio frequency methods like WiFi or radar offer lower spatial resolution \cite{Poppe2010,Lara2013Survey,Lee2022RadarHAR}. LiDAR addresses these limitations by producing 3D point clouds that encode geometry without identifying textures, making it inherently privacy-preserving. As an active sensor, it is also robust to ambient lighting. Its spatial resolution is valuable for recognizing similar human motions, as illustrated in Figure \ref{fig:pc_person_detection}.

Current systems across modalities often rely on large deep models (DL) that process images, skeletons, voxels, or raw points \cite{Poppe2010,haresamudram2025past}. These models can require extensive annotation, significant compute, and offer limited interpretability, which complicates deployment at the edge or in clinical and industrial settings. This motivates compact and transparent representations that work with limited data while preserving discriminative information on posture and motion.

To address this gap, we propose a spectral graph approach to HAR. Each LiDAR frame is converted to a proximity graph where points are nodes and edges link spatial neighbors. From the graph Laplacian we extract its eigenvalues, the “vibrational modes” of the 3D configuration, and statistics of the associated eigenvectors. This spectral information forms a compact “structural fingerprint” of pose per frame. We then compute temporal statistics of these fingerprints over a sliding window to capture motion dynamics, and train standard machine learning (ML) classifiers on the resulting vectors. The Laplacian spectrum preserves a direct link to geometry, which supports interpretability \cite{Shahbaz2023_Connectivity}. The full pipeline is shown in Figure \ref{fig:architecture}. Our contributions include a graph-based representation of LiDAR point cloud sequences for activity recognition, the use of the Laplacian eigen spectrum with eigenvector statistics as a robust descriptor of pose and its temporal evolution, an empirical demonstration on MM Fi that this feature engineering with windowed aggregation attains high accuracy under a strict subject independent protocol and improves over skeleton based baselines, and a modality agnostic framework that can be applied to point clouds from other sensors such as WiFi or radar reconstructions \cite{maatta2025spatio}.

% Our contributions include: 
% \begin{itemize}
% \item A graph-based representation of LiDAR point cloud sequences for human activity recognition.
% \item The use of the graph Laplacian eigen-spectrum as a robust descriptor for human pose and its temporal dynamics.
% \item An empirical demonstration that our feature engineering and temporal aggregation approach achieves high accuracy on a large public dataset, presenting a viable alternative to end-to-end DL methods.
% \item An extensible framework for activity recognition from any modality that produces point clouds, such as WiFi or radar reconstructions \cite{maatta2025spatio}.
% \end{itemize}

%%%%%%%%%%%%%%%%%%%%%%%%%%%%%%%%%%%%%%%%%%%%%%%%%%%%%%%%%%%%%
%
%
%       RELATED WORK
%
%
%%%%%%%%%%%%%%%%%%%%%%%%%%%%%%%%%%%%%%%%%%%%%%%%%%%%%%%%%%%%%

% [TINO NOTES]
% - Potential related work with Radars: https://www.sciencedirect.com/science/article/pii/S2542660524003998

\section{Related Work}

Research in Human Activity Recognition (HAR) has progressed through several paradigms tied to sensing technologies. Early work with wearable sensors established a pipeline of extracting handcrafted statistical or spectral features from accelerometer data for classical machine learning models \cite{Lara2013Survey,haresamudram2025past}. While DL models that learn features from raw signals have become common, the efficiency of classical pipelines remains attractive for edge devices with limited compute. In vision-based HAR, initial methods using handcrafted spatio-temporal descriptors for RGB video \cite{Poppe2010} were largely superseded by end-to-end deep networks, such as 3D CNNs and two-stream architectures that fuse appearance and motion information \cite{gu2021survey}. To address the privacy and illumination challenges of cameras, research shifted to depth sensors, which enabled the extraction of 3D human skeletons. This graph-based abstraction of body joints became a primary input for Graph Neural Networks (GNNs), which excel at modeling dependencies between joints \cite{srimath2021human}. A similar pattern is found in RF sensing, where sparse radar point clouds are often converted into an estimated skeleton before recognition \cite{Lee2022RadarHAR}. Across these advanced modalities, a common pattern is a two-stage pipeline that first estimates a simplified human pose and then classifies the activity.

LiDAR-based HAR has followed a similar trajectory. Early explorations with 2D scanners supported simple tasks like gait tracking \cite{Rinchi2023IoTMag}, while the advent of 3D LiDAR enabled more sophisticated handcrafted features, such as gait energy images from sparse point streams \cite{Benedek2018}. However, the current literature is dominated by DL pipelines that often rely on the same two-stage, pose-then-classify approach. Recent works convert LiDAR data into depth images to extract skeletons for CNN+LSTM models \cite{meng2025indoor} or apply advanced GNNs directly to the derived skeleton graphs \cite{yu2022versatile}. The official HAR baselines for the MM-Fi dataset are a clear example of this trend, applying GNNs like AGCN \cite{Shi2019_Skeleton} and CTRGCN \cite{Chen2021_Skeleton} to skeletons derived from the LiDAR data \cite{Yang2023mm}. This reliance on an intermediate skeleton representation can discard fine-grained geometric detail from the full-body point cloud and propagate errors from the initial pose estimation step.

Our work proposes an alternative by leveraging Spectral Graph Theory (SGT). The concept of the Laplace-Beltrami spectrum as a "Shape-DNA" for describing geometry is well-established for meshes \cite{reuter2009discrete} and has been successfully extended to point clouds \cite{liang2012geometric}. The power of spectral descriptors for 3D shape analysis, even without DL, has been recently demonstrated in tasks like point cloud matching \cite{bastico2024coupled}. Within HAR, however, "spectral" methods almost exclusively refer to GCNs that operate on skeleton graphs \cite{peng2020learning,huang2020part}, not the direct use of the spectrum as a primary feature. The use of Laplacian eigenvalues and eigenvector statistics derived from a graph of a raw point cloud remains an underexplored approach for LiDAR-based HAR. We address this gap by building a proximity graph directly on the segmented person points and deriving features from its Laplacian spectrum, thereby maintaining a direct link to the full-body geometry and avoiding an intermediate pose estimation step \cite{wei2021graph}.

\section{Proposed Methodology}

\begin{figure*}[ht!]
\vspace{-1mm}
 \begin{center}
   \includegraphics*[width=0.98\textwidth]{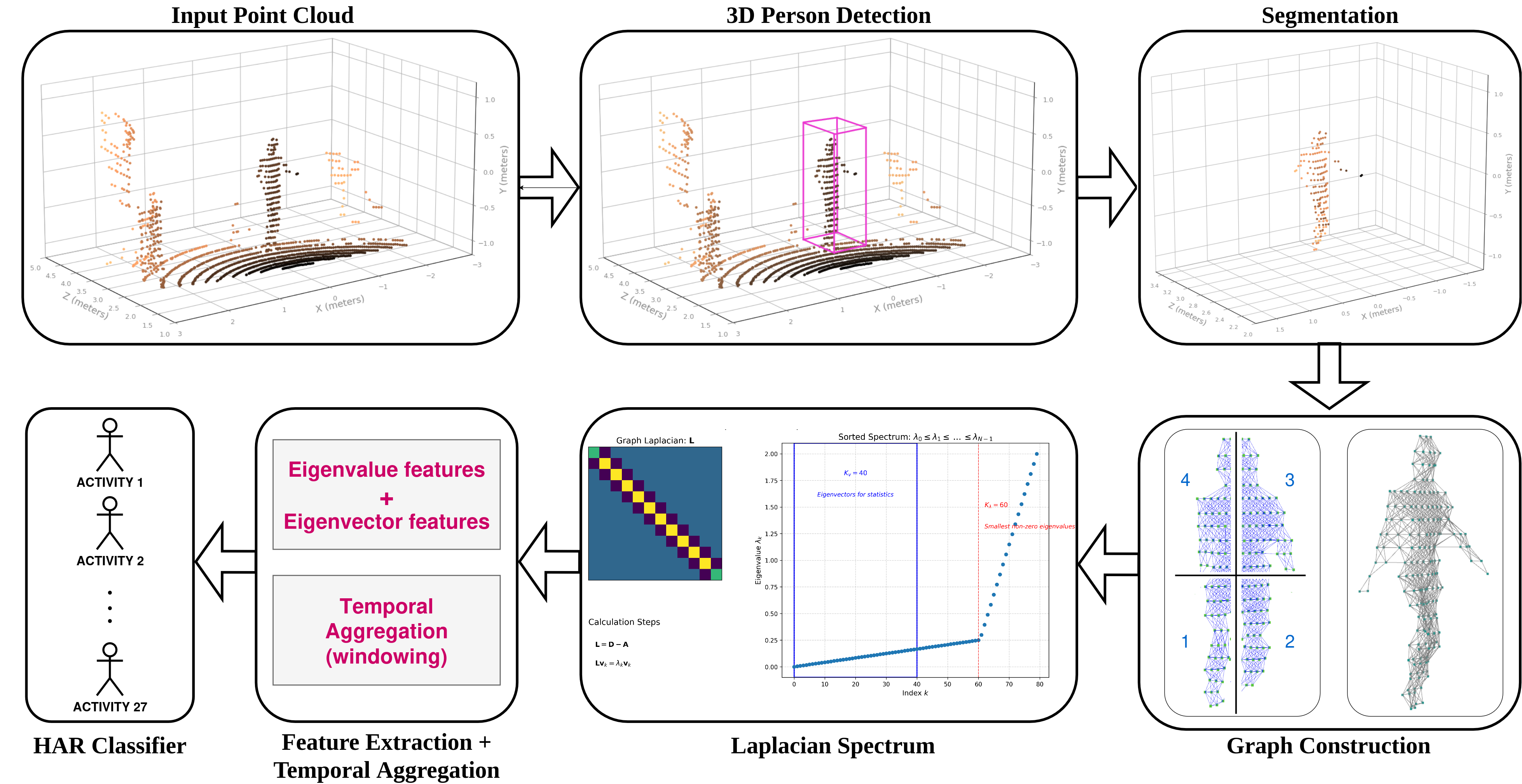}
 \end{center}
 \vspace{-3mm}
 \caption{An overview of the proposed Laplacian-based HAR pipeline. The process starts with a raw input LiDAR frame. A 3D person detector localizes the subject, and these points are segmented from the scene. A proximity graph is constructed from the point cloud of the subject, and divided in 4 quadrants. We then compute the Laplacian matrix and its spectrum from the graph of the whole body and its quadrants. Finally, features derived from the eigenvalues and eigenvectors are aggregated over a temporal window and supplied to a machine learning classifier to predict the human activity.}
 \label{fig:architecture}
 \vspace{-4mm}
\end{figure*}

This work introduces a framework for Human Activity Recognition (HAR) that transforms raw LiDAR point clouds into descriptive feature vectors using graph spectral analysis. The process, illustrated in Figure \ref{fig:architecture}, begins with a raw point cloud from a single frame. A 3D object detection model first identifies and localizes the human subject within the scene, and the points corresponding to the subject are segmented from the background. From this segmented point cloud, a proximity graph is constructed to represent the subject's physical pose. The subsequent stage involves spectral analysis, where the graph's Laplacian matrix is computed to obtain its eigenvalues and eigenvectors. The features derived from this spectrum are then aggregated over a temporal window to capture the dynamics of the activity. Finally, these high-dimensional feature vectors are used to train a machine learning classifier to determine the activity class. The design is invariant to point ordering and tolerant to non-uniform sampling density, and it requires no mesh or skeleton extraction.

% This work introduces a framework for HAR that transforms raw LiDAR point clouds into descriptive feature vectors using graph spectral analysis, as shown in Figure \ref{fig:architecture}. 

\subsection{3D person Detection}  % Tharindu stuff

\begin{figure}[ht!]
\vspace{-1mm}
 \begin{center}
   \includegraphics*[width=0.49\textwidth]{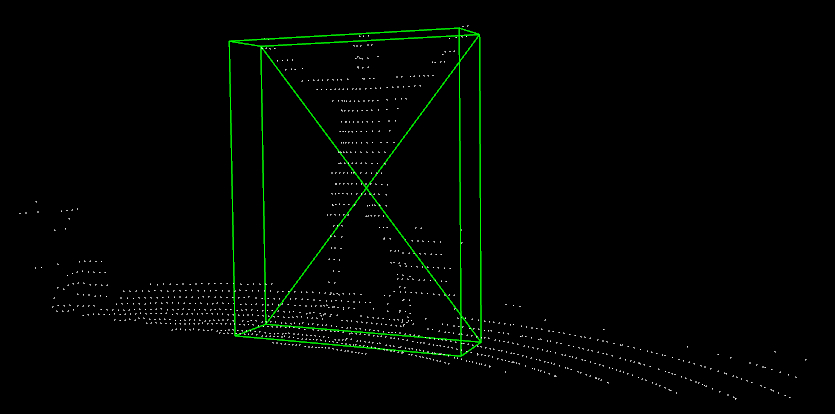}
 \end{center}
 \vspace{-3mm}
 \caption{3D person detection on an indoor LiDAR frame using PV-RCNN. The detector outputs a bounding box around the subject, whose points are then used in the graph-spectral HAR pipeline.}
 \label{fig:3dpersondetection}
 % \vspace{-4mm}
\end{figure}

People detection in point cloud data was implemented using the PVRCNN (PointVoxel-Region-based Convolutional Neural Network) architecture \cite{shi2020pv} within the OpenPCDet framework, specifically configured for single-class person detection, as shown in Figure \ref{fig:3dpersondetection}. The network uses a sophisticated multi-stage architecture comprising a Voxel Feature Encoder (VFE) with MeanVFE, a 3D backbone using VoxelBackBone8x for sparse voxel processing, and a 2D backbone with BaseBEVBackBone for bird's-eye view feature extraction. The model was trained in the MM-Fi dataset \cite{Yang2023mm} using LiDAR point cloud data, with anchor sizes specifically tuned for human detection ($0.8 \times 0.6 \times 1.73$ meters) and optimized hyperparameters including a learning rate of 0.01, a batch size of 8, and training for 80 epochs. Since the MM-Fi dataset lacked ground truth labels for people detection in the form of 3D bounding boxes, the training labels were derived from 3D pose estimation data with coordinate system corrections applied to align the axes properly.

\vspace{-2mm}
\subsection{Point Cloud to Graph Representation}
\label{sec:pointcloud_to_graph}

The segmented point cloud of the subject, denoted as $\mathcal{P} = \{\boldsymbol{p}_i\}_{i=1}^N \subset \mathbb{R}^3$, is transformed into an undirected graph $G=(V, E)$, where the vertices $V$ correspond to the points in $\mathcal{P}$. Edges are formed using a radius graph model, where an edge $(v_i, v_j)$ exists if the Euclidean distance between the corresponding points $\boldsymbol{p}_i$ and $\boldsymbol{p}_j$ is less than a radius $r$:
\begin{equation}
    (v_i, v_j) \in E \iff \|\boldsymbol{p}_i - \boldsymbol{p}_j\|_2 < r.
\end{equation}
The radius $r$, a hyperparameter that defines the scale of local connectivity. Based on empirical evaluation, we set $r=0.15$ meters. To create more discriminative features, we also partition the point cloud P into four spatial quadrants using the median of the lateral and vertical coordinates. A separate graph is constructed for the whole body and for each quadrant, enabling the combination of global and part-based spectral features. This construction converts the unstructured point set into a graph where spatial proximity is encoded by edges, effectively capturing the underlying geometry of the human pose.

\vspace{-2mm}
\subsection{Laplacian Spectral Analysis}
\label{sec:spectral_analysis}
To analyze the structure encoded within each graph $G$, we utilize Spectral Graph Theory \cite{wei2021graph}. For a graph $G$ with adjacency $A$ and diagonal degree matrix $D$ with $D_{ii}=\deg(v_i)$, the combinatorial Laplacian is \(L = D - A\).  The Laplacian matrix $L$ is a symmetric, positive semi-definite matrix whose spectral properties reveal intrinsic structural information about the graph. Solving the eigenproblem for $L$, given by \(L\boldsymbol{u}_i = \lambda_i \boldsymbol{u}_i,\),  yields non-negative eigenvalues $0=\lambda_0 \le \lambda_1 \le \cdots \le \lambda_{N-1}$ and orthonormal eigenvectors $\{\boldsymbol{u}_0, \boldsymbol{u}_1, \dots, \boldsymbol{u}_{N-1}\}$, where each $\boldsymbol{u}_i \in \mathbb{R}^N$. The set $\{\lambda_k\}$ is the \textit{graph spectrum}. It is permutation-invariant and summarizes connectivity and geometry. Changes in human pose modify the edges and thus the spectrum. Eigenvalues provide a compact description of shape and connectivity, while eigenvectors describe spatial “modes” over the point cloud.

\vspace{-2mm}
\subsection{Spatio-Temporal Feature Engineering}
\label{sec:feature_extraction}

To transform the raw spectral data into a format suitable for machine learning, we designed a multi-stage feature engineering pipeline. This process first converts the spectrum of each frame into a per-frame feature vector and then aggregates these vectors over a temporal window to capture motion dynamics.

\vspace{-3mm}
\subsubsection{Per-Frame Feature Extraction.}
\vspace{-1mm}
For each frame $t$, we derive features from its eigenvalues and eigenvectors. For the Eigenvalue Features ($f^{\text{val}}_t$), we explore two methods. The first, a \textit{Statistical Summary}, computes a fixed-size vector of 6 statistics (mean, standard deviation, median, 25th/75th percentiles, max) over the non-trivial spectrum ($\lambda_1, \dots, \lambda_{N-1}$), capturing the overall distribution of the spectral values. The second, a \textit{Direct Selection} method, uses the first $K_{\text{val}}$ eigenvalues as a raw feature vector, which describes the low-frequency signature of the pose. For the Eigenvector Features ($f^{\text{vec}}_t$), we extract statistics from the first $K_{\text{vec}}$ non-trivial eigenvectors ($\boldsymbol{u}_1, \dots, \boldsymbol{u}_{K_{\text{vec}}}$). For each eigenvector $\boldsymbol{u}_i$, we compute a set of 7 statistics (mean, standard deviation, max, min, kurtosis, Shannon entropy of its squared components, and the range of its absolute values), which are then concatenated.

\vspace{-3mm}
\subsubsection{Temporal Window Aggregation}
\vspace{-1mm}
To incorporate temporal context, we use a sliding window of size $W$ frames with a stride of $S$. For each window, the sequence of per-frame features is aggregated into a single vector using one of two primary strategies. The first, \textit{Temporal Statistics}, computes a rich set of statistics (e.g., mean, standard deviation, range, median, interquartile range, skewness, kurtosis) for each feature dimension across all $W$ frames. This method captures the central tendency, variability, and dynamics of the features over time. The second strategy, \textit{Concatenation}, simply concatenates the per-frame feature vectors into one long vector, preserving the exact temporal sequence but resulting in a higher-dimensional representation.

% ($\sim$4000 features)

\vspace{-3mm}
\subsubsection{Final Feature Vector}
\vspace{-1mm}
% Combining the two per-frame eigenvalue methods with the two aggregation methods yields four experimental strategies: (i) \textbf{Strategy A} combines the Statistical Summary with Temporal Statistics, (ii) \textbf{Strategy B} combines Direct Selection with Temporal Statistics, (iii) \textbf{Strategy C} combines the \textit{Statistical Summary} with Concatenation, and (iv) \textbf{Strategy D} combines \textit{Direct Selection} with Concatenation. The eigenvector features are always aggregated using the Temporal Statistics method, resulting in a single vector $F^{\text{vec}}_{\text{window}}$. The final feature vector for each window is the concatenation of the eigenvalue and eigenvector components:
% \begin{equation}
%     F_{\text{window}} = [ F^{\text{val}}_{\text{window}} \, ; \, F^{\text{vec}}_{\text{window}} ].
% \end{equation}

Combining the two eigenvalue options with two aggregation choices yields four eigenvalue strategies:
\begin{enumerate}
    \item Strategy A computes the six value summary per frame and applies temporal aggregation to obtain \(F^{\text{val}}_{\!A}\).
    \item Strategy B keeps the first \(K_{\text{val}}\) eigenvalues per frame and applies temporal aggregation to obtain \(F^{\text{val}}_{\!B}\).
    \item Strategy C concatenates the per frame six value summaries across the window without temporal statistics, yielding \(F^{\text{val}}_{\!C}=[\,f^{\text{stat}}_{1};\dots;f^{\text{stat}}_{W}\,]\).
    \item Strategy D concatenates the per frame selected eigenvalues across the window, yielding \(F^{\text{val}}_{\!D}=[\,f^{\text{sel}}_{1};\dots;f^{\text{sel}}_{W}\,]\). 
\end{enumerate}

Eigenvector features are always temporally aggregated, producing \(F^{\text{vec}}_{\text{window}}\).

When spatial partitions are enabled, the same procedure is applied to the whole body and to predefined parts (four quadrants from median splits on the lateral and vertical axes and a lower body mask). The final feature vector for each window is the concatenation of the eigenvalue and eigenvector components (if utilized): 
\begin{equation}
    F_{\text{window}} = [ F^{\text{val}}_{\text{window}} \, ; \, F^{\text{vec}}_{\text{window}} ].
\end{equation}
This design provides a compact spectral summary of pose per frame, captures its temporal evolution within a window, and supports whole body or part based representations through simple concatenation.

\vspace{-2mm}
\subsection{Classification}
\label{sec:classification}

We evaluate the engineered features using a strict subject-independent protocol, where subjects in the test set are entirely unseen during training. Before classification, features are scaled fitted only on the training data. Optional dimensionality reduction with Principal Component Analysis (PCA) is also explored. We assess performance with several classifiers: a Support Vector Machine (SVM), a Random Forest (RF), and a Multi-layer Perceptron (MLP).

%%%%%%%%%%%%%%%%%%%%%%%%%%%%%%%%%%%%%%%%%%%%%%%%%%%%%%%%%%%%%
%
%
%       EXPERIMENTAL SETUP
%
%
%%%%%%%%%%%%%%%%%%%%%%%%%%%%%%%%%%%%%%%%%%%%%%%%%%%%%%%%%%%%%

\section{Evaluation Methodology}

\subsection{Benchmark database and Evaluation Protocol}

We use the MM\textendash Fi dataset \cite{Yang2023mm}, which contains recordings of 40 participants (11 female, 29 male) performing 27 actions, split into 14 daily activities and 13 rehabilitation exercises. The dataset provides six sensing modalities; in this work we use only the LiDAR point clouds and the action annotations. The LiDAR sensor is an Ouster OS1 with 32 channels, operating at 10 Hz. Data were captured in four environments covering two room layouts of $8.5\,\text{m}\times 7.8\,\text{m}$, with 10 subjects per environment.

We adopt the MM\mbox{-}Fi \textit{cross\_scene\_split} \cite{Yang2023mm} to test generalization to new subjects and environments. Training uses 30 subjects recorded in E1–E3, and testing uses the remaining 10 subjects recorded only in E4. This split ensures that both subjects and the room are unseen during training.

  \vspace{-2mm}
\subsection{Performance Metrics}

We report window\mbox{-}level accuracy as the primary metric, together with macro F1 and balanced accuracy (unweighted means of per\mbox{-}class F1 and recall, respectively). For interpretability we include a confusion matrix with absolute counts and row\mbox{-}normalized percentages. Each metric is given as mean $\pm$ standard deviation. The mean is computed directly over the $N$ test windows. The standard deviation is estimated by nonparametric bootstrap on windows: draw $N$ windows with replacement, recompute the metric, and repeat $B{=}1000$ times. The reported uncertainty is the empirical standard deviation across the $B$ bootstrap estimates. We also report Top\mbox{-}1 and Top\mbox{-}5 accuracies following the MM\mbox{-}Fi evaluation protocol for ranked hypotheses, computed at the window level.

\vspace{-2mm}
\subsection{Implementation Details}

For each frame, we build an $\varepsilon$-graph with radius $r{=}0.15$ m unless stated otherwise, compute the combinatorial Laplacian $L{=}D{-}A$, and obtain eigenpairs sorted by ascending eigenvalue. Temporal aggregation uses sliding windows of $w\in\{2,4,6\}$\,s with a stride of $0.5$ s, which at $10$ Hz corresponds to $W\in\{20,40,60\}$ frames and a stride of $5$ frames. Unless noted, spectral parameters are $K_{\text{val}}{=}60$ and $K_{\text{vec}}{=}40$. When spatial partitions are enabled we compute descriptors for the whole body (part=0), four quadrants (parts $1$–$4$), and the lower body (part=5), then concatenate the window vectors. Before classification, features are standardized with \texttt{StandardScaler}. We use Random Forest, SVM, and MLP models with library default hyperparameters unless explicitly stated otherwise in a given experiment. PCA is off by default. The implementation is based on scikit-learn and PyTorch (Python 3.10).

%%%%%%%%%%%%%%%%%%%%%%%%%%%%%%%%%%%%%%%%%%%%%%%%%%%%%%%%%%%%%
%
%
%       EXPERIMENTAL RESULTS
%
%
%%%%%%%%%%%%%%%%%%%%%%%%%%%%%%%%%%%%%%%%%%%%%%%%%%%%%%%%%%%%%

\section{Experimental Results}

This section reports results for the proposed LiDAR based HAR. We first compare eigenvalue feature strategies across window durations on the 13 and 27 class tasks. We then study spatial partitions and compare with MM-Fi skeleton based baselines.

% , graph radius, and runtime,

% \textcolor{red}{
% \begin{itemize}
%     \item Include ablation with different combinations of the body parts...
%     \item Measure time consumption of feature extraction for a window + Inference
%     \item Add the baseline result in the MM-Fi database. Take a look to the Top-1 and Top-5 Metrics.
%     \item Ablation study about radious in the graph
% \end{itemize}
% }
\vspace{-2mm}
\subsection{Impact of Feature Strategy and Window Size}

We evaluated the impact of our feature engineering choices, with results presented in Table~\ref{tab:acc_strategies_windows}. The analysis reveals two clear findings. First, \textbf{Strategy B}, which computes temporal statistics on the first $K_{\text{val}}=60$ selected eigenvalues, consistently outperforms all other strategies. Second, augmenting any strategy with eigenvector features provides a significant accuracy improvement, which confirms that statistics describing the spatial distribution of the graphs primary modes offer complementary information to the eigenvalues alone.

% Table~\ref{tab:acc_strategies_windows} reports window level accuracy for the four eigenvalue strategies (A–D) on the 13 and 27 class tasks with window lengths $w\in\{2,4,6\}$\,s at 10 Hz.

\begin{table*}[ht!]
\def\arraystretch{1.05}
\setlength{\tabcolsep}{1.2em} % Adjust column spacing
\centering
\caption{Window level accuracy (\%) under the subject split for four eigenvalue strategies across 13 and 27 activities and window durations  $w\in\{2,4,6\}$\,s. Each cell shows mean $\pm$ bootstrap STD over test windows. Window stride is 0.5 seconds. Classifier is Random Forest. Radius $r=0.15$ m. In bold the best result per strategy and window length.}
\label{tab:acc_strategies_windows}
% \vspace{-2mm}
% \resizebox{\linewidth}{!}{%
\scalebox{0.9}{%
\begin{tabular}{|l|c|c|c|c|c|c|}
\hline
& \multicolumn{3}{|c|}{13 activities} & \multicolumn{3}{|c|}{27 activities} \\
\cline{2-4} \cline{5-7}
Strategy & w=2 & w=4 & w=6 & w=2 & w=4 & w=6 \\
\hline
A & 76.89 ± 0.80 & 81.45 ± 0.80 & 82.96 ± 0.91 & 73.10 ± 0.60 & 78.06 ± 0.62 & 81.91 ± 0.64 \\
A (+ $f^{\text{vec}}_t$) & 84.36 ± 0.71 &  87.42 ± 0.71 & 88.05 ± 0.81 & 74.51 ± 0.57 & 78.13 ± 0.61 & 79.91 ± 0.66 \\
B ($K_{\text{val}}=60$) & 82.71 ± 0.71 & 88.96 ± 0.66 & 89.29 ± 0.75 & 75.86 ± 0.58 & 83.14 ± 0.55 &  83.70 ± 0.65 \\
B ($K_{\text{val}}=60$ + $f^{\text{vec}}_t$) & \textbf{88.64 ± 0.59} & \textbf{91.81 ± 0.61} & \textbf{90.41 ± 0.72} & \textbf{81.31 ± 0.53} &  \textbf{83.59 ± 0.57} &  \textbf{84.47 ± 0.61} \\
C  & 65.82 ± 0.87 & 66.33 ± 0.97 & 66.09 ± 1.15 &  52.89 ± 0.66 & 52.96 ± 0.73 & 53.11 ± 0.83 \\
C (+ $f^{\text{vec}}_t$) & 80.99 ± 0.75 & 84.48 ± 0.77 & 84.08 ± 0.86 & 66.60 ± 0.65 & 69.83 ± 0.69 & 72.54 ± 0.73 \\
D ($K_{\text{val}}=60$) & 74.62 ± 0.85 & 73.89 ± 0.96 & 70.18 ± 1.08 & 66.86 ± 0.64 &  66.97 ± 0.73 & 66.07 ± 0.80 \\
D ($K_{\text{val}}=60$ + $f^{\text{vec}}_t$) & 83.74 ± 0.71 & 86.33 ± 0.72 & 80.47 ± 0.94 & 69.03 ± 0.62 & 73.46 ± 0.63 & 74.23 ± 0.69 \\

\hline
\end{tabular}}
% \vspace{-4mm}
\end{table*}

The optimal window duration is task-dependent. For the 13-class rehabilitation set, a 4-second window is optimal (91.81\% accuracy), a duration that aligns with the short and repetitive nature of these exercises. For the more challenging 27-class set, a longer 6-second window achieves a slightly better accuracy of 84.47\%, suggesting the extended temporal context is beneficial for distinguishing between finer-grained activities (e.g., daily activities such as picking up things). Based on this analysis, we identify Strategy B with eigenvector features as our best configuration for further evaluation.

% Strategy B is consistently best across all window lengths and both label sets. Windows of 
% $w=4$ s give the highest accuracy, which aligns with the dataset where most activities are short and repetitive and a full motion cycle fits within 4 s. Adding eigenvector features (+ $F^{\text{vec}}_{\text{window}}$) further improves accuracy because these statistics capture how low frequency spatial modes distribute over the body, complement the eigenvalues, and are stable to sign flips and point ordering. Following this comparison, we analyze the best configuration in more detail. Figure~\ref{fig:cm_class} shows the confusion matrices for the 13 class and 27 class tasks. The 13 class matrix presents a sharper diagonal, while the 27 class matrix shows increased off-diagonal mass due to finer class granularity. The effect of $w=4$ s + $F^{\text{vec}}_{\text{window}}$ is visible as stronger diagonal entries and fewer confusions among closely related activities.

\begin{figure}[ht!]
% \vspace{-1mm}
 \begin{center}
   \includegraphics*[width=0.49\textwidth]{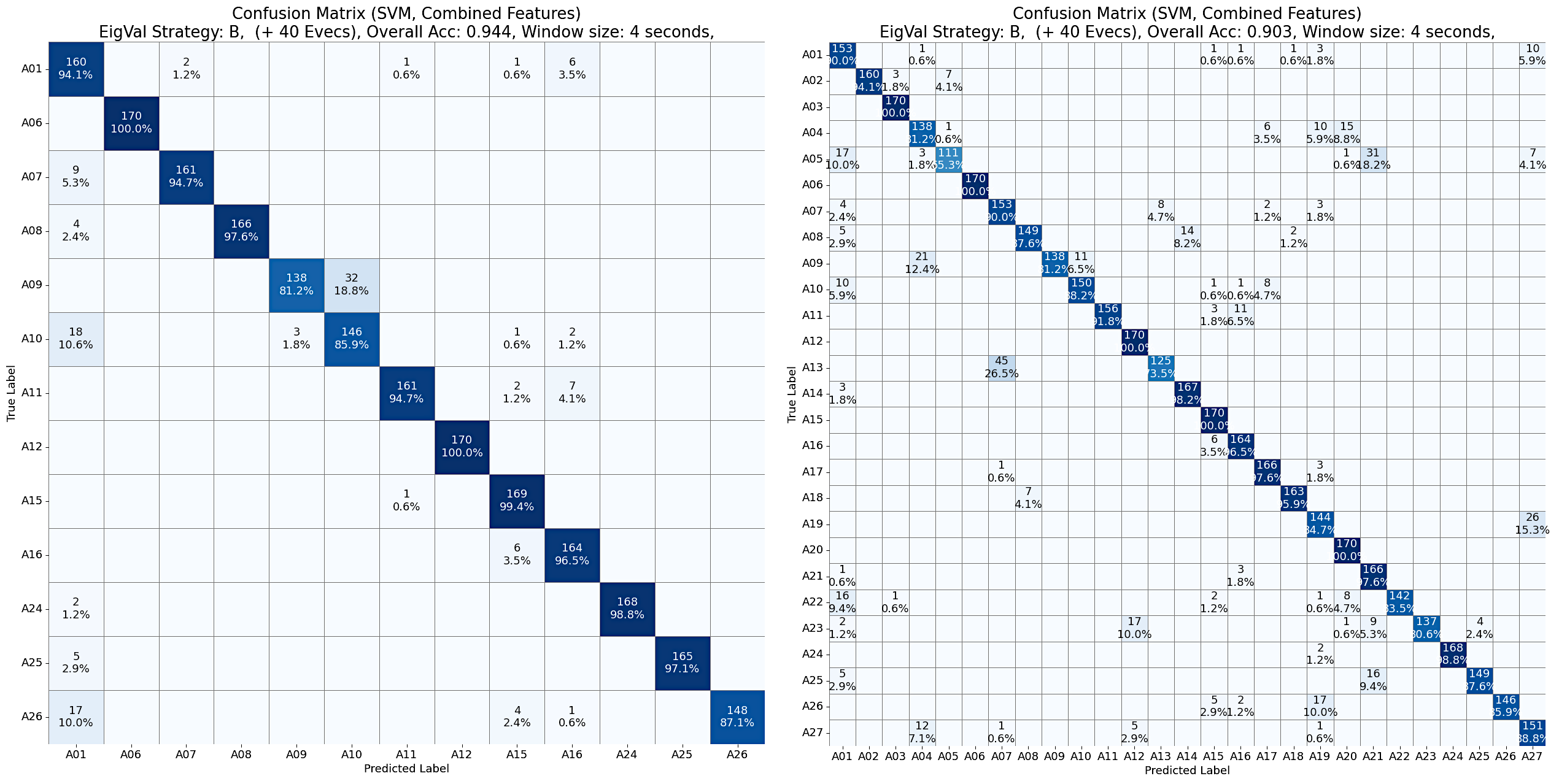}
 \end{center}
 \vspace{-6mm}
 \caption{Confusion matrices for 13 and 27 activities with the best setting (SVM, $r=0.15$ m, standardized features, whole body + quadrants).}
 \label{fig:cm_class}
 % \vspace{-4mm}
\end{figure}

Following this analysis, we examine the detailed performance of our best model configuration. Figure~\ref{fig:cm_class} presents the confusion matrices for both the 13-class and 27-class activity sets, obtained using an SVM classifier with a 4-second window. For the 13-class task (left), the model achieves an overall accuracy of 94.4\%. The matrix shows a strong diagonal, with most rehabilitation exercises like ``Squat'' (A12) and ``Lunge (toward left)'' (A15) being classified with nearly perfect accuracy. Some notable confusions occur between activities with similar overall body dynamics, such as  ``Lunge (toward left-front)'' (A09) and ``Lunge (toward right-front)'' (A10).

% \vspace{-2mm}
\subsection{Ablation Study on Global and Local Details}

% An ablation study on spatial partitions, with results in Table~\ref{tab:quadrants_classifiers}, was conducted to analyze the contribution of global and local features. The results show that features derived solely from the whole body are insufficient, while features from the four spatial quadrants alone provide a strong baseline, indicating that limb dynamics are highly discriminative.

% To understand the contributions of global and local geometric information, we conducted an ablation study by constructing features from different spatial partitions of the point cloud. The results, presented in Table~\ref{tab:quadrants_classifiers}, show that relying solely on features from the whole body or isolated parts such as the lower body produces poor classification accuracy. In contrast, using only the four spatial quadrants provides a strong baseline, which indicates that the dynamics of individual limbs are highly discriminative for these activities.

Table~\ref{tab:quadrants_classifiers} evaluates spatial partitions to separate global posture from local limb dynamics. The weakest configuration is using only the legs (part 5), which yields the lowest accuracy on both 13 and 27 classes. Many actions in MM-Fi are dominated by upper body motion or coordinated upper–lower movements (rehabilitation), so restricting descriptors to the lower body discards critical variance. Using only the whole body (part 0) also underperforms, likely because global aggregation smooths out localized, high-frequency limb patterns that distinguish similar activities.

% To understand the contributions of global and local geometric information, we conducted an ablation study by constructing features from different spatial partitions of the point cloud. The best configuration is the combination of whole body and four quadrants (0 + 1–4) in Table~\ref{tab:quadrants_classifiers}. It mixes global motion cues from the full body with fine details from local regions, so repetitive limb dynamics are captured without losing overall trajectory and posture. Among classifiers, SVM gives the highest accuracy for both 13 and 27 classes. We reported Random Forest in the Table \ref{tab:acc_strategies_windows} because it is a strong and stable baseline that is less sensitive to feature scaling and hyperparameter choices, offers straightforward feature importance analysis, and gives consistent trends across window lengths and strategies.

In contrast, the four-quadrant representation (parts 1–4) is already strong. Partitioning by lateral and vertical medians preserves the local dynamics of arms and legs and reduces the cancelation effects present in whole-body pooling. Adding a partial global context further helps: three-quadrant combinations improve over single partitions. The best results arise from combining whole body with all four quadrants (0 + 1–4), which joins overall trajectory and posture with detailed limb motion. This synergy is most visible for repetitive actions, where global stability disambiguates similar limb trajectories, while quadrants retain discriminative detail.

\begin{table}[ht!]
% \vspace{-2mm}
\def\arraystretch{1.05}
\setlength{\tabcolsep}{0.3em}
\centering
\caption{Window–level recognition accuracy (\%) under the subject split for different spatial partitions and classifiers. }
\label{tab:quadrants_classifiers}
\resizebox{\linewidth}{!}{%
\begin{tabular}{|l|ccc|ccc|}
\hline
& \multicolumn{3}{|c|}{13 activities} & \multicolumn{3}{|c|}{27 activities} \\
\cline{2-4} \cline{5-7}
Combination & RF & SVM & MLP & RF & SVM & MLP \\
\hline
Whole body only (part=0)                 & 54.98 & 61.04 & 59.05 & 41.48 & 53.14 & 51.68 \\
Four quadrants only (parts 1–4)          & 88.14 & 92.81 & 92.67 & 79.78 & 87.71 & 78.91 \\
Three quadrants only (parts 3–5)         & 86.02 & 90.05 & 91.95 &  82.72 & 86.82 & 83.42 \\
Whole body + four quadrants (0 + 1–4)    & 91.81 & \textbf{94.39} & 93.57 & 83.88 & \textbf{90.33} & 84.38 \\
Whole body + upper quadrants (0 + 3,4)   & 87.83 & 92.26 & 92.76 & 82.48 & 87.58 & 84.90 \\
Only legs (part=5)                       & 51.45 & 60.81 & 61.40 &  42.31 & 50.26 & 49.83 \\
\hline
\end{tabular}}

\vspace{2pt}
\footnotesize\textit{Notes:} Vector features extracted using Strategy B + Eigenvector features. Window length = 4 seconds. PCA=True (n\_components=1500)
% \vspace{-6mm}
\end{table}

Across classifiers, the SVM achieves the highest accuracy with the combined representation, reaching 94.39\% for 13 activities and 90.33\% for 27 activities. We used Random Forest in Table~\ref{tab:acc_strategies_windows} because it is a strong and stable baseline that is less sensitive to feature scaling and hyperparameters, offers straightforward feature importance analysis, and yields consistent trends across window sizes and strategies.

% The highest performance is achieved by combining features from the whole body with those from the four quadrants. This synergy is effective because it integrates global information on overall posture and trajectory with fine-grained details of limb movements captured by the feature-based features. Among the tested classifiers, the SVM consistently achieves the best results with this combined feature set, reaching a peak accuracy of 94.39\% for the 13-class task and 90.33\% for the 27-class task.

\subsection{Ablation Study on Graph Radius Impact}

Table~\ref{tab:quadrants_classifiers2} compares two radius thresholds for the $\varepsilon$-graph, $r{=}0.15$ m, and $r{=}0.20$ m, on the 13-class task with an SVM. Increasing the radius consistently improves accuracy across window lengths and spatial partitions. The gain is most pronounced for configurations that pool over large regions, for example, the whole body $w{=}4$ rises from $61.04$ to $66.15$ (+5.11 pp) and legs only from $60.81$ to $66.47$ (+5.66 pp). Four quadrants also benefit, though with smaller margins, from $92.81$ to $94.25$ (+1.44 pp) at $w{=}4$. The best result is obtained by combining whole body with four quadrants, which reaches $95.88$ at $r{=}0.20$ and $w{=}4$, slightly higher than the $w{=}6$ setting. These trends indicate that $r{=}0.20$ produces a better connected graph, stabilizes the Laplacian spectrum, and reduces fragmentation in sparse frames. Very small radii can produce disconnected components and noisy eigenvectors, while excessively large radii would risk oversmoothing. In this range, $r{=}0.20$ strikes a good balance.

\begin{table}[ht!]
% \vspace{-2mm}
\def\arraystretch{1.05}
\setlength{\tabcolsep}{0.3em}
\centering
\caption{Window–level recognition accuracy (\%) under the subject split for different spatial partitions and classifiers and graph radio construction. Results for 13 activity classification. SVM classifier}
\label{tab:quadrants_classifiers2}
\resizebox{\linewidth}{!}{%
\begin{tabular}{|l|ccc|ccc|}
\hline
& \multicolumn{3}{|c|}{$r=0.15$ m.} & \multicolumn{3}{|c|}{$r=0.20$ m.} \\
\cline{2-4} \cline{5-7}
Combination & w=2 & w=4 & w=6 & w=2 & w=4 & w=6 \\
\hline
Whole body only (part=0)                 & 58.35 & 61.04 & 60.89    & 64.58 & 66.15 & 65.86 \\
Four quadrants only (parts 1–4)          & 90.70 & 92.81 & 92.43    & 93.04 & 94.25 & 94.14 \\
Three quadrants only (parts 3–5)         & 88.64 & 90.05 & 89.29    & 92.56 & 93.08 & 93.55 \\
Whole body + four quadrants              & 93.22 & 94.39 & \textbf{94.85}  & 95.27 & \textbf{95.88} & 95.68 \\
Whole body + upper quadrants             & 90.48 & 92.26 & 92.72    & 93.55 & 94.57 & 95.15 \\
Only legs (part=5)                       & 56.96 & 60.81 & 60.47    & 64.18 & 66.47 & 64.44 \\
\hline
\end{tabular}}

\vspace{2pt}
\footnotesize\textit{Notes:} Vector features extracted using Strategy B + Eigenvector features. SVM Classifier (kernel='rbf', C=1.0). PCA=True (n\_components=1500)
% \vspace{-6mm}
\end{table}

\begin{figure}[ht!]
% \vspace{-1mm}
 \begin{center}
   \includegraphics*[width=0.49\textwidth]{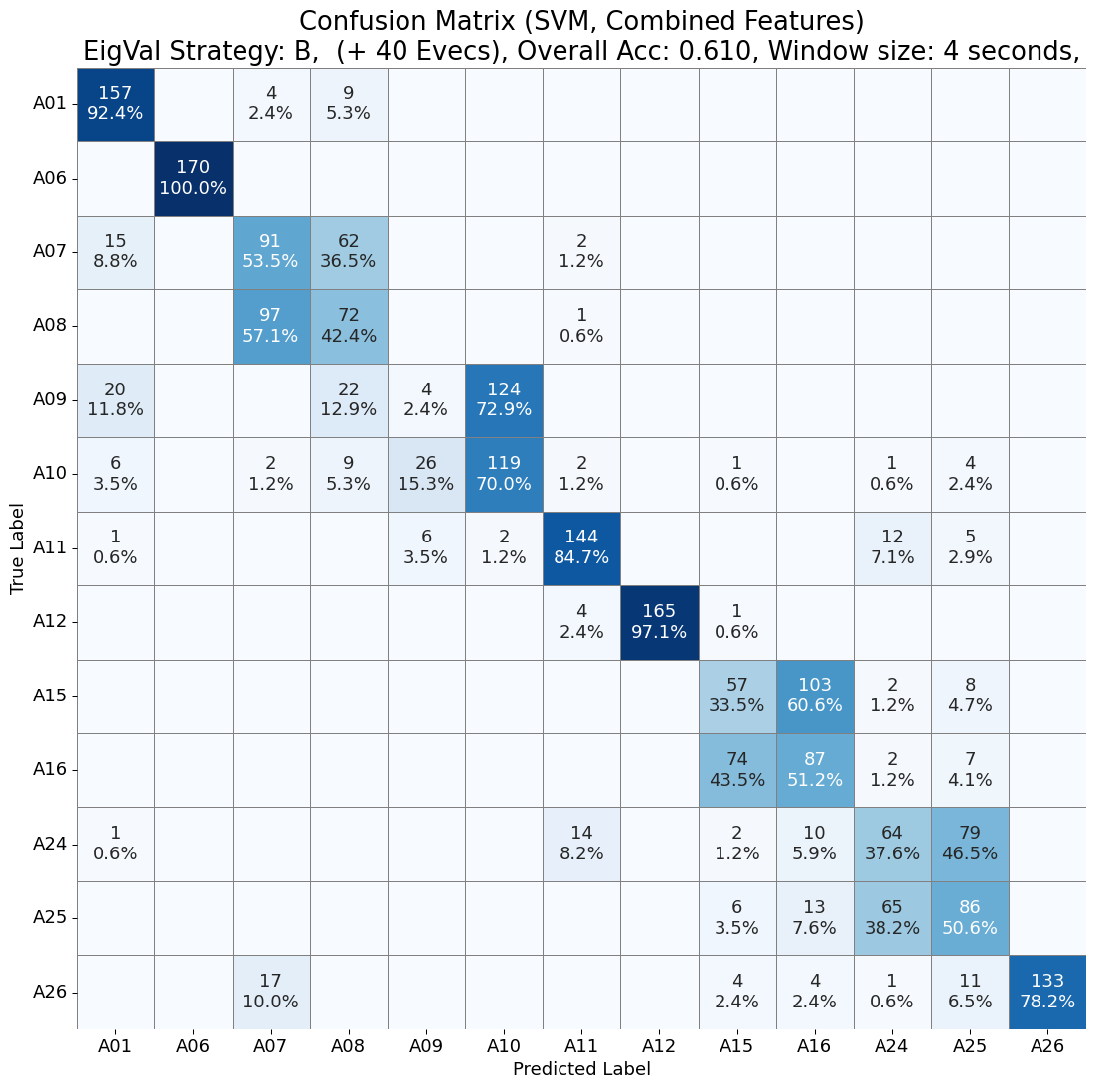}
 \end{center}
 \vspace{-4mm}
 \caption{Confusion matrix for 13 activities using only whole body features (SVM, $r=0.15$ m, $w=4$ s, standardized features). Left and right symmetric classes are frequently confused, for example A07 vs A08, A15 vs A16, and A24 vs A25, because the undirected Laplacian and whole body pooling do not encode direction.}
 \label{fig:cm_wholebody}
 % \vspace{-4mm}
\end{figure}

Figure~\ref{fig:cm_wholebody} shows a confusion matrix for the whole body only representation at $r{=}0.15$ m and $w{=}4$ s, which is the weakest configuration in Table~\ref{tab:quadrants_classifiers2}. Clear confusions appear between symmetric pairs such as A07 vs A08 (limb extension left vs right), A15 vs A16 (lunge toward left vs right), and A24 vs A25 (body extension left vs right). An undirected Laplacian and whole body pooling do not encode direction, and eigenvectors are sign ambiguous, so left and right variants produce near identical spectra. Adding lateral quadrants breaks this symmetry and recovers side-specific cues, which explains the strong improvement when combining the whole body with four quadrants.

% \vspace{-2mm}
\subsection{Comparison with State-of-the-Art on the MM-Fi Dataset}

We evaluated our method against the LiDAR-based results of two skeleton-based action recognition models, AGCN \cite{Shi2019_Skeleton} and CTRGCN \cite{Chen2021_Skeleton}, using the benchmarks from the original MM-Fi publication \cite{Yang2023mm}. These benchmark methods operate in a two-stage pipeline by first estimating a 3D skeleton from the point cloud and then classifying the action from that skeleton. Our approach, in contrast, computes features directly from the raw point cloud geometry. To provide a comprehensive comparison, we assessed our method under two evaluation protocols. First, to create a direct and fair comparison, we adopted the protocol used by the benchmark methods. This involves training and testing exclusively on subjects within the E04 environment, partitioned into a 7-subject set for training and a 3-subject set for testing. Second, we applied our own stricter protocol to assess generalization more rigorously, training on 30 subjects from environments E01, E02, and E03 and testing on 10 different subjects in the unseen E04 environment. For both protocols, our method used a 30-frame window and our best-performing feature configuration (Strategy B with eigenvector features and a graph radius of r = 0.15).

Table \ref{tab:comparison_sota} presents the results for the 27-activity classification task. Under the identical E04 (7/3 split) cross-validation protocol, our method achieves a Top-1 accuracy of 81.26\%, substantially outperforming both AGCN (54.44\%) and CTRGCN (35.97\%). When trained on more data and tested under our stricter protocol, the performance of our model increases further to 90.33\%.

% We compare against LiDAR-based skeleton pipelines AGCN \cite{Shi2019_Skeleton} and CTRGCN \cite{Chen2021_Skeleton} from \cite{Yang2023mm}. Those methods first estimate a 3D skeleton and then classify actions. Our method operates directly on point clouds. We report two protocols: the MM-Fi E04 split with 7 training and 3 test subjects, and a stricter cross-scene split with 30 subjects from E01–E03 for training and 10 subjects from E04 for testing. In both cases, we use 30-frame windows, SVM classifier, Strategy B with eigenvector features, and radius $r=0.15$ m. The results for 27 classes are shown in Table~\ref{tab:comparison_sota}. Our method improves Top-1 over the baselines on the E04 split and further improves under the cross-scene protocol.

\begin{table}[ht!]
\vspace{-5mm}
\def\arraystretch{1.1}
\setlength{\tabcolsep}{0.4em}
\centering
\caption{Top-1 and Top-5 accuracy (\%) on MM-Fi LiDAR for 27 activities.}
\vspace{1mm}
\label{tab:comparison_sota}
\resizebox{\linewidth}{!}{%
\begin{tabular}{|l|l|c|c|}
\hline
\textbf{Method} & \textbf{Protocol (Train/Test Subjects)} & \textbf{Top 1 (\%)} & \textbf{Top 5 (\%)} \\
\hline
AGCN~\cite{Shi2019_Skeleton} & Env. E04 (7/3) & 54.44 & 91.03 \\
CTRGCN~\cite{Chen2021_Skeleton} & Env. E04 (7/3) & 35.97 & 74.12 \\
\hline
\textbf{Ours (SVM)} & Env. E04 (7/3) & \textbf{81.26} & \textbf{96.66} \\
\textbf{Ours (SVM)} & Env. E01-E03 (30) / E04 (10) & \textbf{90.33} & \textbf{97.82} \\
\hline
\end{tabular}}
% \vspace{3pt}
% \footnotesize\textit{Notes:} The results of AGCN~\cite{Shi2019_Skeleton} and CTRGCN~\cite{Chen2021_Skeleton} were from~\cite{Yang2023mm}
\end{table}

The superior performance in the direct comparison highlights the effectiveness of our feature representation. By operating in a complete point cloud, our method avoids the information loss inherent in abstracting the human form into a simplified skeleton and is not affected by potential errors from an intermediate pose estimation step. The further performance improvement under the stricter protocol suggests that our method also scales well with more training data and generalizes effectively to both unseen subjects and new environments.

% The superior performance in the direct comparison highlights the effectiveness of our feature representation. By operating on the complete point cloud, our method avoids the information loss inherent in abstracting the human form into a simplified skeleton and is not affected by potential errors from an intermediate pose estimation step. The further performance improvement under the stricter protocol suggests that our method also scales well with more training data and generalizes effectively to both unseen subjects and new environments.

%%%%%%%%%%%%%%%%%%%%%%%%%%%%%%%%%%%%%%%%%%%%%%%%%%%%%%%%%%%%%
%
%
%       CONCLUSION
%
%
%%%%%%%%%%%%%%%%%%%%%%%%%%%%%%%%%%%%%%%%%%%%%%%%%%%%%%%%%%%%%
\vspace{-6mm}
\section{Conclusion}

We presented a LiDAR-based HAR method that operates directly on point clouds. Each frame is mapped to an $\varepsilon$ graph, the combinatorial Laplacian is computed, and the eigenvalues together with the eigenvector statistics define the descriptors per frame. Temporal statistics over sliding windows produce fixed length vectors that are classified with standard models. Person segmentation is obtained with PV-RCNN, so no skeleton extraction is required. In MM Fi the method reached 94.39\% accuracy on the 13 rehabilitation classes and 90.33\% on the 27 class set under a strict cross-scenario subject split. With $r{=}0.20$ m and $w{=}4$ s the 13 class result rose to 95.88\%. The study showed that Strategy B with eigenvectors, a 4 s window, and the fusion of whole body with four quadrants give the best trade-off between global posture and local limb dynamics. The comparison against MM-Fi skeleton baselines under the Top-k protocol confirmed the benefit of operating on the full point cloud. These findings indicate that the Laplacian spectrum provides a compact structural fingerprint of pose and motion that is accurate, interpretable, and practical for point cloud HAR. Future work will extend the evaluation to additional datasets, study the radius of the graph over a broader range and adaptive choices, and analyze the feature space to reduce the dimensionality. In Strategy B with eigenvectors the representation is 4400 dimensions per body part, which motivates feature selection, compact projections, or learned embeddings. We also plan to profile runtime on embedded hardware and to explore multipart fusion with temporal models while keeping the same spectral front-end.

% We presented a LiDAR-based HAR method that operates directly on point clouds. Each frame is modeled as an $\varepsilon$-graph, the combinatorial Laplacian is computed, and the eigenvalue together with the eigenvector statistics form per frame descriptors. Temporal statistics over sliding windows produce fixed length vectors that are classified with standard models. Person segmentation is obtained with PV-RCNN, so no skeleton extraction is required. On MM-Fi the method achieved 94.39\% accuracy on 13 rehabilitation classes and 90.33\% on the full 27 class set under a strict cross scene subject split, and it outperformed LiDAR skeleton baselines under the MM-Fi Top-k protocol. These results indicate that the Laplacian spectrum provides a structural fingerprint of pose and motion that is accurate, interpretable, and practical for point cloud HAR. Future work will extend the evaluation to additional datasets, study the radius of the graph over a broader range and adaptive choices, and analyze the feature space to reduce the dimensionality. In Strategy B with eigenvectors the representation is 4400 dimensions per body part, which motivates feature selection, compact projections, or learned embeddings. We also plan to profile runtime on embedded hardware and to investigate multi part fusion with temporal models while keeping the same spectral front-end.

%Remove it now, we will add it back in the camera ready in the right format. We save space.
% [TINO COMMENT] ---> Miguel put here the Converge Grant number.
\section*{Acknowledgment}
The research was supported by the Business Finland WISEC project (Grant 3630/31/2024), the University of Oulu and the Research Council of Finland (former Academy of Finland) 6G Flagship Programme (Grant Number: $346208$), Profi5 HiDyn programme (326291), and Profi7 Hybrid intelligence program (352788). The authors acknowledge the CSC-IT Center for Science, Finland, for computational resources (Graph generation, feature extraction, and model training were performed on the Mahti supercomputer (CSC Finnish Supercomputer Center), which is a BullSequana XH2000 system. The model training did use only CPUs.

% the Horizon Europe CONVERGE project (Grant 1010948)

\bibliographystyle{IEEEbib}
\bibliography{strings,refs}

\begin{thebibliography}{10}

\bibitem{Rinchi2023IoTMag}
Omar Rinchi, Hakim Ghazzai, Ahmad Alsharoa, and Yehia Massoud,
\newblock ``Lidar technology for human activity recognition: Outlooks and challenges,''
\newblock {\em IEEE Internet of Things Magazine}, vol. 6, no. 2, pp. 143--150, 2023.

\bibitem{Poppe2010}
Ronald Poppe,
\newblock ``A survey on vision-based human action recognition,''
\newblock {\em Image and Vision Computing}, 2010.

\bibitem{Lara2013Survey}
Oscar~D. Lara and Miguel~A. Labrador,
\newblock ``A survey on human activity recognition using wearable sensors,''
\newblock {\em IEEE Communications Surveys \& Tutorials}, 2013.

\bibitem{Lee2022RadarHAR}
Gawon Lee and Jihie Kim,
\newblock ``Improving human activity recognition for sparse radar point clouds: A graph neural network model with pre-trained 3d human-joint coordinates,''
\newblock {\em Applied Sciences}, vol. 12, no. 4, pp. 2168, 2022.

\bibitem{haresamudram2025past}
Harish Haresamudram, Chi~Ian Tang, Sungho Suh, Paul Lukowicz, and Thomas Ploetz,
\newblock ``Past, present, and future of sensor-based human activity recognition using wearables: A surveying tutorial on a still challenging task,''
\newblock {\em Proceedings of the ACM on Interactive, Mobile, Wearable and Ubiquitous Technologies}, vol. 9, no. 2, pp. 1--44, 2025.

\bibitem{Shahbaz2023_Connectivity}
Karim Shahbaz, Madhu~N. Belur, and Ajay Ganesh,
\newblock ``{Algebraic Connectivity: Local and Global Maximizer Graphs},''
\newblock {\em IEEE Transactions on Network Science and Engineering}, 2023.

\bibitem{maatta2025spatio}
Tuomas M{\"a}{\"a}tt{\"a}, Sasan Sharifipour, Miguel~Bordallo L{\'o}pez, and Constantino~{\'A}lvarez Casado,
\newblock ``Spatio-temporal 3d point clouds from wi-fi-csi data via transformer networks,''
\newblock in {\em 2025 IEEE 5th International Symposium on Joint Communications \& Sensing (JC\&S)}. IEEE, 2025, pp. 1--6.

\bibitem{gu2021survey}
Fuqiang Gu, Mu-Huan Chung, Mark Chignell, Shahrokh Valaee, Baoding Zhou, and Xue Liu,
\newblock ``A survey on deep learning for human activity recognition,''
\newblock {\em ACM Computing Surveys (CSUR)}, vol. 54, no. 8, pp. 1--34, 2021.

\bibitem{srimath2021human}
Sivanvita Srimath, Yang Ye, Krishanu Sarker, Rajshekhar Sunderraman, and Shihao Ji,
\newblock ``Human activity recognition from rgb video streams using 1d-cnns,''
\newblock in {\em IEEE SmartWorld, ubiquitous intelligence \& computing, advanced \& trusted computing, scalable computing \& communications, internet of people and smart city innovation}. IEEE, 2021.

\bibitem{Benedek2018}
Csaba Benedek, Bence G\'{a}lai, Bal\'{a}zs Nagy, and Zsolt Jank\'{o},
\newblock ``Lidar-based gait analysis and activity recognition in a 4d surveillance system,''
\newblock {\em IEEE Transactions on Circuits and Systems for Video Technology}, vol. 28, no. 1, pp. 101--113, 2018.

\bibitem{meng2025indoor}
Xiang Meng, Mondher Bouazizi, Zhaojie Li, and Tomoaki Ohtsuki,
\newblock ``Indoor human activity recognition using multiple dynamic nonlinear mapping applied to 3d lidar-collected data,''
\newblock {\em IEEE Internet of Things Journal}, 2025.

\bibitem{yu2022versatile}
Jiahui Yu, Yingke Xu, Hang Chen, and Zhaojie Ju,
\newblock ``Versatile graph neural networks toward intuitive human activity understanding,''
\newblock {\em IEEE Transactions on Neural Networks and Learning Systems}, vol. 35, no. 7, pp. 8869--8881, 2022.

\bibitem{Shi2019_Skeleton}
Lei Shi, Yifan Zhang, Jian Cheng, and Hanqing Lu,
\newblock ``Two-stream adaptive graph convolutional networks for skeleton-based action recognition,''
\newblock in {\em IEEE/CVF Conference on Computer Vision and Pattern Recognition}, 2019.

\bibitem{Chen2021_Skeleton}
Yuxin Chen, Ziqi Zhang, Chunfeng Yuan, Bing Li, Ying Deng, and Weiming Hu,
\newblock ``Channel-wise topology refinement graph convolution for skeleton-based action recognition,''
\newblock in {\em IEEE/CVF Conference on Computer Vision and Pattern Recognition}, 2021.

\bibitem{Yang2023mm}
Jianfei Yang, He~Huang, Yunjiao Zhou, Xinyan Chen, Yuecong Xu, Shangguan Yuan, Hanhua Zou, Chunxiao Lu, and Lihua Xie,
\newblock ``{MM-Fi}: Multi-modal non-intrusive 4d human dataset for versatile wireless sensing,''
\newblock in {\em Advances in Neural Information Processing Systems (NeurIPS)}, 2023.

\bibitem{reuter2009discrete}
Martin Reuter, Silvia Biasotti, Daniela Giorgi, Giuseppe Patan{\`e}, and Michela Spagnuolo,
\newblock ``Discrete laplace--beltrami operators for shape analysis and segmentation,''
\newblock {\em Computers \& Graphics}, vol. 33, no. 3, pp. 381--390, 2009.

\bibitem{liang2012geometric}
Jian Liang, Rongjie Lai, Tsz~Wai Wong, and Hongkai Zhao,
\newblock ``Geometric understanding of point clouds using laplace-beltrami operator,''
\newblock in {\em 2012 IEEE conference on computer vision and pattern recognition}. IEEE, 2012, pp. 214--221.

\bibitem{bastico2024coupled}
Matteo Bastico, Etienne Decenci{\`e}re, Laurent Cort{\'e}, Yannick Tillier, and David Ryckelynck,
\newblock ``Coupled laplacian eigenmaps for locally-aware 3d rigid point cloud matching,''
\newblock in {\em Proceedings of the IEEE/CVF Conference on Computer Vision and Pattern Recognition}, 2024, pp. 3447--3458.

\bibitem{peng2020learning}
Wei Peng, Xiaopeng Hong, Haoyu Chen, and Guoying Zhao,
\newblock ``Learning graph convolutional network for skeleton-based human action recognition by neural searching,''
\newblock in {\em AAAI conference on artificial intelligence}, 2020.

\bibitem{huang2020part}
Linjiang Huang, Yan Huang, Wanli Ouyang, and Liang Wang,
\newblock ``Part-level graph convolutional network for skeleton-based action recognition,''
\newblock in {\em Proceedings of the AAAI conference on artificial intelligence}, 2020, vol.~34, pp. 11045--11052.

\bibitem{wei2021graph}
H~Wei,
\newblock ``Graph spectral point cloud processing,''
\newblock in {\em Graph Spectral Image Processing}. 2021.

\bibitem{shi2020pv}
Shaoshuai Shi, Chaoxu Guo, Li~Jiang, Zhe Wang, Jianping Shi, Xiaogang Wang, and Hongsheng Li,
\newblock ``Pv-rcnn: Point-voxel feature set abstraction for 3d object detection,''
\newblock in {\em IEEE Conference on Computer Vision and Pattern Recognition}, 2020.

\end{thebibliography}

\end{document}